\documentclass[letterpaper, 10 pt, conference]{ieeeconf}  

\IEEEoverridecommandlockouts                              

\usepackage{multirow}
\usepackage{booktabs}
\usepackage{amssymb}  
\usepackage{graphicx}
\usepackage{amsmath}
\usepackage{siunitx}

\usepackage{marvosym}  

\title{
\large \bf
HRO: Hierarchical ``Room-to-Object'' Framework for Zero-Shot Object Goal Navigation with Large Language Models$\dagger$
}

\author{Luyuan Jia$^{1,2,3}$, and Yinfeng Yu$^{1,2,3}$$^{,\mbox{\Letter}}$%
\thanks{\small $\dagger$This research was financially supported by the National Natural Science Foundation of China (Grant No. 62463029).}
\thanks{\small $^1$School of Computer Science and Technology, Xinjiang University, Urumqi 830017, China.}
\thanks{\small $^2$Joint Research Laboratory for Embodied Intelligence, Xinjiang University.}
\thanks{\small $^3$Joint International Research Laboratory of Silk Road Multilingual Cognitive Computing, Xinjiang University.}
\thanks{\small $^{\mbox{\Letter}}$Yinfeng Yu is the corresponding author (Email: yuyinfeng@xju.edu.cn).}%
}

\begin{document}

\maketitle
\thispagestyle{empty}
\pagestyle{empty}

\begin{abstract}

Zero-shot object-goal navigation aims to enable an intelligent agent to explore and navigate to objects of unknown categories in an unfamiliar environment without specific target training. In zero-shot navigation tasks, pre-trained large models are usually employed to leverage their prior knowledge for guiding the agent's navigation. However, existing zero-shot object-goal navigation methods based on large language models (LLMs) merely utilize LLMs as ``flat'' reasoning tools to directly associate objects or regions. They lack the hierarchical spatial cognition modeling of human-like room semantics to object localization, which leads to strong blindness in exploration, insufficient accuracy in semantic association, and failure to fully unleash the common-sense reasoning potential of LLMs. This paper proposes an LLM-driven hierarchical room-to-object (HRO) framework for zero-shot object-goal navigation, which guides the agent to explore and navigate to the target object in a coarse-to-fine manner. Experiments on Gibson and HM3D datasets verify that our HRO framework achieves superior success rate and generalization over existing LLM-based methods, underscoring LLMs’ strong potential for zero-shot object-goal navigation.

\end{abstract}

\section{INTRODUCTION}

Different from other artificial intelligence tasks~\cite{YinfengICLR2022saavn,yu2023measuring,yang2026beyond,yu2025dope,yu2025dgfnet,fu2025fsdenet,mattursun2024bss,zhang2024nonlinear,li2025audio,zhang2025advancing,zhang2025iterative,yu2025dynamic,wang2025modality,cao2024vnet}, object goal navigation in unknown environments is critical for home service robots and autonomous exploration. Existing LLM-based zero-shot methods, such as ESC~\cite{zhou2023esc} and L3MVN~\cite{yu2023l3mvn}, only perform flat reasoning to directly associate objects and regions. They lack human-like hierarchical spatial cognition: humans first judge room type from observed objects, then infer target locations based on room functionality. Without such modeling, exploration is blind, semantic association is imprecise, and LLM commonsense potential is underutilized.
\begin{figure}[t]
\centering
\includegraphics[width=0.95\linewidth]{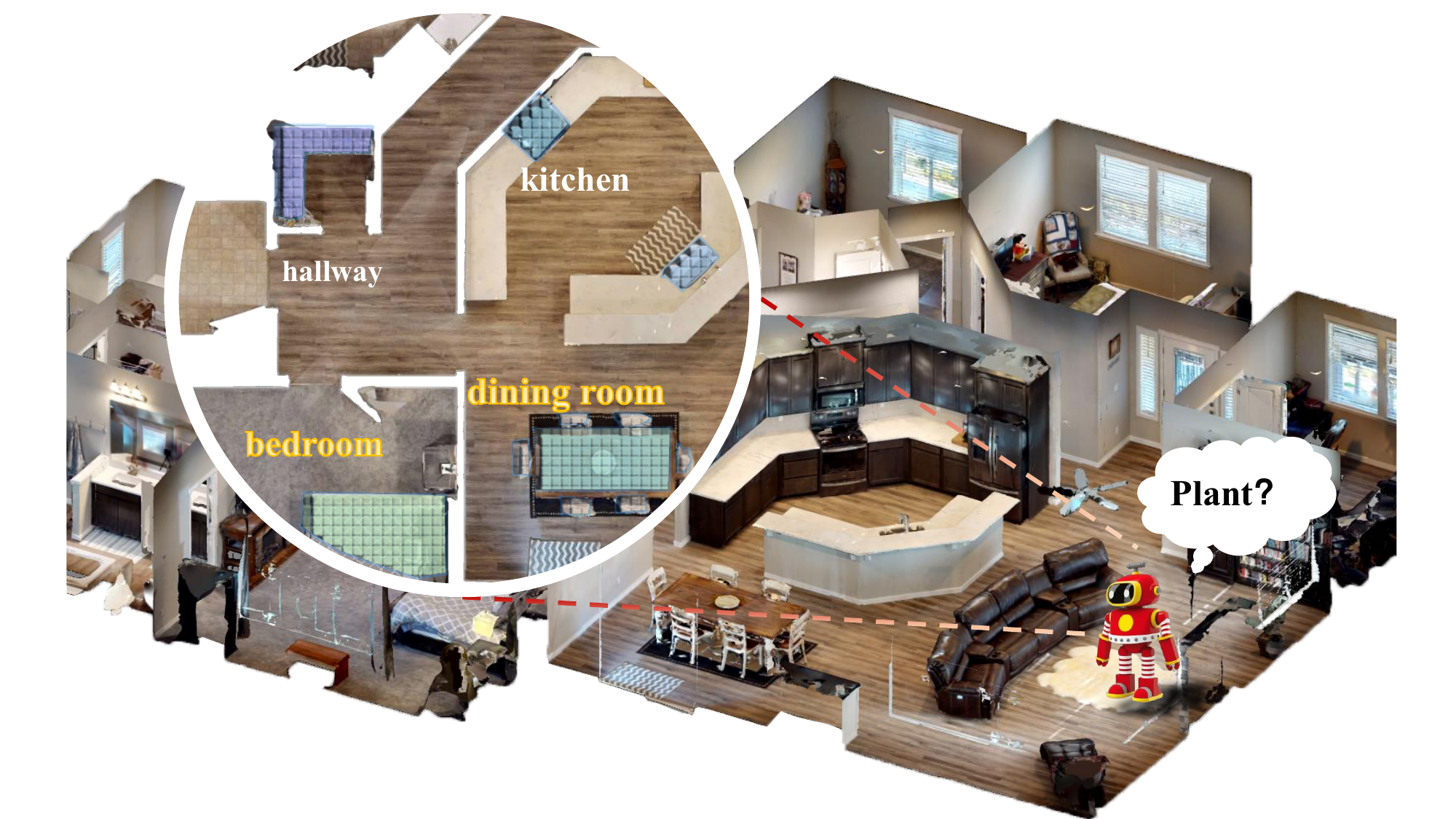}
\caption{The figure illustrates the HRO framework. 
Taking ``finding plant'' as an example, the agent infers room semantics from observed objects and selects target regions using object-room commonsense, embodying the hierarchical ``object to room to target'' reasoning mechanism.
} \label{fig1}
\end{figure}

To address this, we propose HRO, a hierarchical LLM-driven zero-shot navigation framework. Following human cognition, HRO adopts a coarse-to-fine paradigm and decomposes navigation into three modules: (1) The high-level module uses an LLM to infer room type distributions from observed objects; (2) The mid-level module evaluates candidate frontiers via target-oriented semantic affinity scoring; (3) The low-level module plans the shortest path and executes actions.

This framework uses ``room type'' as a core semantic bridge between LLM common sense and object localization. A concurrent work, LROGNav ~\cite{SUN2025103135}, also explores room–object relations but relies on supervised training, while HRO is training-free and hierarchical. It aligns agent exploration with the human logic of ``first judging the scenario, then finding the object'' and improves semantic guidance. As shown in Fig.~\ref{fig1}, taking ``finding plant'' as an example, the agent first infers room semantics from observed objects, then searches regions with high object-room affinity, achieving hierarchical perception-to-target navigation.

\begin{figure*}[t]
\centering 
\includegraphics[width=0.7\textwidth]{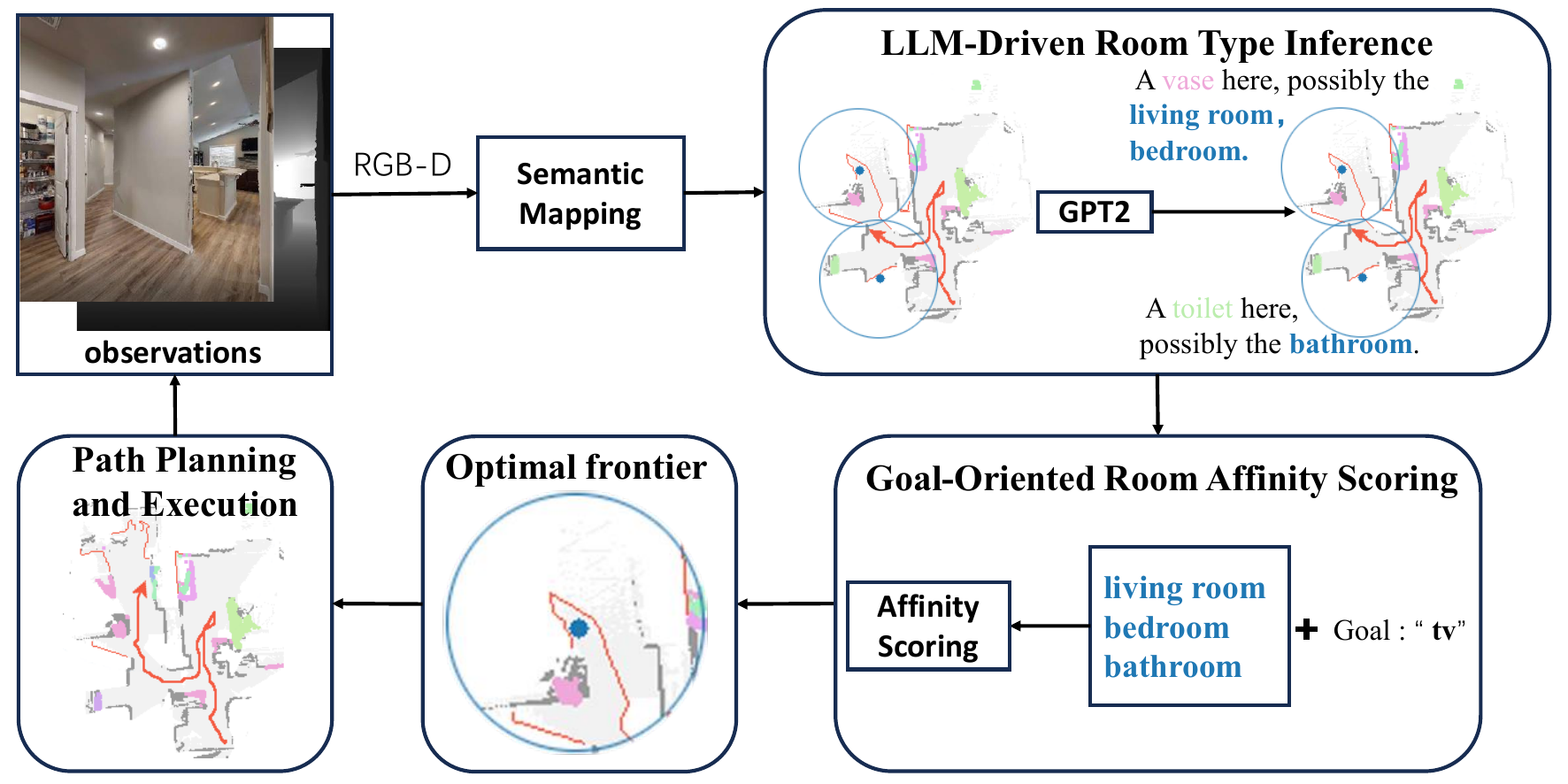}
\caption{The architecture of the HRO Framework. This framework demonstrates a hierarchical ``room-to-object'' navigation process. 
} \label{fig2}
\end{figure*}

Our contributions are summarized as follows:
\begin{itemize}
    \item We propose HRO, a hierarchical ``room-to-object'' zero-shot framework with a three-tier structure (room reasoning, region decision, action execution), where ``room type'' serves as the core semantic bridge to enhance exploration pertinence.
    \item Relying on the rich prior knowledge of LLMs, our framework can directly adapt to unseen objects and unknown scenes without additional annotation of object and scene training data or task-specific fine-tuning, providing a new solution for zero-shot object navigation.
    \item Our method achieves the best success rate on two navigation benchmark datasets, and its performance outperforms baseline methods. This verifies the great potential of LLMs in the development of the zero-shot object-goal navigation field.
\end{itemize}

\section{Related Works}
Traditional object navigation methods rely on extensive annotated data, but pre-trained large models now enable zero-shot learning with strong generalization~\cite{long2024instructnav,sakamoto2024map,patel2025get}. Some VLM-based methods (e.g., VLFM~\cite{yokoyama2024vlfm}, GAMap~\cite{yuan2024gamap}) perform frontier selection for zero-shot navigation. We focus on LLM-based methods for semantic reasoning potential.

Existing LLM-based methods including L3MVN~\cite{yu2023l3mvn}, ESC~\cite{zhou2023esc}, L-ZSON~\cite{dorbala2023can}, and PixNav~\cite{cai2024bridging} adopt flat reasoning to directly associate objects and regions. They lack hierarchical modeling of room–object relations, which limits semantic reasoning accuracy and exploration efficiency. Our hierarchical mechanism fills this gap and provides a more human-like cognitive paradigm for navigation.

Concurrently, LROGNav~\cite{SUN2025103135} utilizes LLM-based room–object relationships for navigation, but relies on supervised learning and labeled room segmentation maps. In contrast, our HRO is zero-shot, training-free, and hierarchical without any annotations or fine-tuning. It performs on-the-fly LLM reasoning via a three-stage pipeline that aligns with human spatial cognition.

\section{The Proposed Method}
\subsection{Problem Formulation}
We focus on zero-shot object-goal navigation, where an agent explores an unknown indoor environment $S_i$ to find a target category $C_i$ without task-specific training, fine-tuning, or prior exposure to $S_i$ and $C_i$. An episode is defined as $E_i = \{ S_i, C_i, P_0 \}$, where $P_0$ denotes the agent's initial pose. At each step $t$, the agent obtains RGB-D observations, position $(x_t, y_t)$, and orientation $\theta_t$. The action space includes move forward, turn left/right, look up/down, and stop. The agent moves \SI{25}{\centi\meter} per step and rotates \SI{30}{\degree} per action. The task succeeds if the agent stops within 1 meter of the target within $T_{\text{max}} = 500$ steps.

\subsection{Overview}
This paper proposes HRO, a hierarchical framework based on LLMs for zero-shot object-goal navigation. The overall workflow is illustrated in Fig.~\ref{fig2}, where the framework addresses challenges by decomposing the zero-shot object navigation problem into three hierarchical levels: The high-level reasoning module infers room types based on an LLM; The mid-level decision module selects the optimal frontier through target-oriented room semantic affinity scoring; The low-level execution module plans the optimal path from the current position to the target region and executes motion commands. Through a coarse-to-fine hierarchical decision-making logic, the framework deeply integrates the high-level semantic reasoning capability of LLMs with low-level geometric path planning. All modules collaborate to form a closed loop. This design enables the agent to ``first judge scene semantics, then conduct targeted object exploration'' like humans in zero-shot scenarios, significantly enhancing the purposefulness and efficiency of navigation.

\subsection{Map Representation}
We first construct the semantic map, drawing on the method proposed in SemExp~\cite{chaplot2020object}. Initially, we initialize a matrix of dimension $K \times M \times M$ to store semantic map information, where $M \times M$ represents the size of the map, and $K$ denotes the total number of channels in the semantic map with $K = C + 2$. The first two channels of the semantic map record obstacle distribution and explored region information, respectively, while the remaining $C$ channels each correspond to a semantic category, used to represent the distribution of different objects in the environment.
In the agent initialization phase, we set its starting position at the center of the map, i.e., the coordinate $(M/2, M/2, 0.0)$. At each time step, the agent acquires current observation data via sensors, including RGB images and depth information. We first extract depth data from the observations and convert it into a 3D point cloud based on camera intrinsic parameters. Subsequently, we use the semantic segmentation model Mask R-CNN~\cite{he2017mask} to predict semantic labels for each point in the point cloud, thereby obtaining a semantic point cloud with category information.
To construct a 2D map representation suitable for navigation, we project the semantic point cloud into a 3D voxel space to form a voxelized representation. By performing vertical projection and fusion of voxels within different height ranges, a 2D semantic map is obtained. Finally, we generate an obstacle map, an explored region map, and semantic maps corresponding to different semantic channels, respectively.

\subsection{LLM-Driven Room Type Inference}
The high-level reasoning module helps the agent better understand its surrounding environment and infers the semantic categories of rooms that may be contained in the current candidate frontiers based on the types of observed objects. It leverages the common-sense knowledge embedded in LLMs as a zero-shot room classifier, eliminating the need to acquire room type annotation information during the training process and simulating humans’ ability to reason about environmental context.

First, we extract the obstacle map and the explored map from the first two channels of the semantic map. To further prevent the agent from colliding with obstacles, we perform a dilation operation on the edges of the obstacle map. We then subtract the dilated obstacle map from the explored map to obtain initial frontier cells. For these frontier cells, we conduct connectivity analysis, perform clustering, and remove clusters of small sizes. Finally, we filter the frontiers using the Mixed Criteria Scoring method mentioned in ~\cite{6095018}—a weighted scoring mechanism that incorporates distance cost and the area occupied by frontier points, defined as follows:
\begin{equation}
S_{f_i}^{MC} = \alpha \cdot A(f_i) + \beta \cdot D(f_i),\ f_i \in F (i = 1,2,\dots,n).
\end{equation}

Where $\alpha$ and $\beta$ are weights, $A(f_{i})$ denotes the area occupied by the frontier, and $D(f_{i})$ represents the shortest path distance from the center point of the frontier to the agent. Through this, we can obtain the candidate frontiers $f_{i} \in F$ at the agent’s current position, and the number of candidate frontiers is set to $n = 4$ here.

We predefine a candidate room type set containing seven common indoor categories: 

\textit{R=\{living room, bedroom, bathroom, kitchen, dining room, office, hallway\}.}

This set of room types covers most functional spaces in indoor environments and supports room semantic reasoning in home settings. By basing our approach on commonsense spatial classification rather than data-driven clustering, we ensure that the zero-shot generalization capability can be extended to other indoor environments in the future.

To enable the LLM to perform the room prediction task, for each candidate frontier, we aggregate known objects within the local window of the frontier and construct a prompt template as follows:
\begin{equation}
\begin{aligned}
\mathrm{Prompt}(O, r) = &\ \mbox{``A room containing ''} + \mathrm{concat}(O) \\
                         &+ \mbox{`` is likely a ''} + r + \mbox{``.''} .
\end{aligned}
\end{equation}

Where $O = \left\{ o_1, o_2, \ldots, o_n \right\}$ represents the types of objects within the local window of the frontier, and $r \in R$ denotes a candidate room type.

For each candidate room type $r_i$, we use the LLM to calculate a compatibility score:
\begin{equation}
\text{score}(r_i \mid O) = -\mathcal{L}_{\text{LLM}}(\text{Prompt}(O, r_i)).
\end{equation}

Where $-\mathcal{L}_{\text{LLM}}$ denotes the language modeling loss of the prompt sequence. Since a lower loss indicates a higher probability, we take its negative value as the frontier room type score. Finally, the room type probability distribution is obtained through softmax normalization:
\begin{equation}
P(r_i \mid O) = \frac{\exp(\text{score}(r_i \mid O))}{\sum\limits_{j=1}^{|R|} \exp(\text{score}(r_j \mid O))}.
\end{equation}

The local window mechanism for frontiers is as follows:
\begin{equation}
O_f = \{ o \in O \mid \text{frontier}(o) \in \text{Window}(f, w, h) \}.
\end{equation}

Where $O = \{o_1, o_2, \ldots, o_n\}$ represents the set of all detected objects in the semantic map, and $\text{Window}(f, w, h)$ defines a search region centered at the frontier point $f$ with a size of $w \times h$.

\begin{figure*}[t]
\includegraphics[width=1\textwidth]{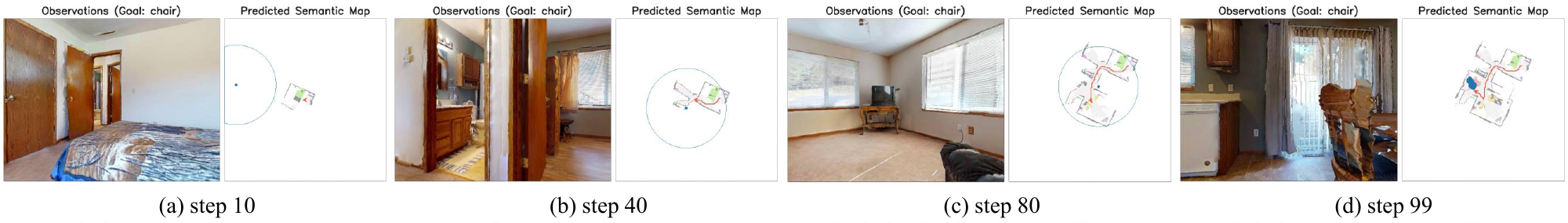}
\caption{The object goal navigation experiment process for finding a chair on the Habitat platform. 
} \label{fig4}
\end{figure*}

\subsection{Goal-Oriented Semantic Affinity Scoring}
The mid-level decision module is responsible for evaluating the potential value of all candidate frontiers in achieving the current navigation goal. Guided by the semantic information from the high-level reasoning module and geometric data, the agent comprehensively assesses and identifies the optimal frontier for exploration. Based on the probability distribution of room types, the mid-level decision module models the affinity between the target object and room types, converting abstract spatial semantics into specific navigation utility.

We constructed a commonsense knowledge base that encodes typical object-room co-occurrence patterns:

\begin{equation}
M:G\to 2^{R}.
\end{equation}

Where $G$ denotes the set of target objects, and the mapping $M$ associates each target $g$ with the set of room types where it is most likely to appear. For instance $M(\text{bed})=\{\text{bedroom}\}$, $M(\text{tv})=\{\text{living room, bedroom}\}$.

If a frontier point leads to any room type where the target object typically appears, this point is considered promising. The affinity score quantifies the degree of matching between the observed objects in the frontier region and the room types that may contain the target object, ensuring that strong evidence for any relevant room type results in a high score. For a given target object $g$ and room probability distribution $P(r|O)$, we calculate the affinity score of the candidate frontier as follows:

\begin{equation}
S_{f_{i}}^{\text{Affinity}}(g, O)=\max _{r\in M(g)}P(r\mid O).
\end{equation}

\subsection{Path Planning and Execution}
The final priority of a candidate frontier is the weighted sum of its semantic affinity score and geometric score, with the formula given as follows:

\begin{equation}
S_{f_{i}}=\alpha\frac{S_{f_{i}}^{\text{Affinity}}(g, O)}{\max _{f}S_{f_{i}}^{\text{Affinity}}(g, O)}+(1-\alpha)S_{f_{i}}^{MC}.
\end{equation}

Based on the reasoning result $S_{f_{i}}$, the agent selects the frontier $f_{i}$ that is most likely to contain the target for exploration. After identifying the frontier with the highest score, the agent uses the Fast Marching Method ~\cite{sethian1996fast} to plan the shortest and optimal path from its current position to the identified frontier. Subsequently, the agent generates the corresponding action $a\in A$ and navigates along this path. The agent incrementally constructs the environmental map through continuous observations and executes low-level actions such as obstacle avoidance based on this map. This deterministic local strategy efficiently guides the agent to perform low-level actions for obstacle avoidance and movement. We repeat this process until the agent achieves the target or the task is terminated.

\section{Experiments}
\subsection{Datasets}
Our experiments are based on two 3D scanned datasets of real-world scenarios: HM3D~\cite{ramakrishnan2021hm3d} and Gibson~\cite{xia2018gibson}. For the Gibson dataset, we selected 1,000 episodes from 5 validation set scenes in the Gibson tiny split and adopted the semantic annotations provided by Armeni et al.~\cite{armeni20193d}. The HM3D dataset has been adapted to the Habitat format, and its validation set includes 2,000 episodes across 20 scenes, covering 6 object categories, namely chair, sofa, potted plant, bed, toilet, and TV. These categories are consistent with those defined in SemExp~\cite{chaplot2020object}.

\subsection{Experiment Details}
Our experiments are based on Habitat-Sim, a high-performance 3D simulator, and Habitat-API, a modular high-level library, both of which belong to the Habitat~\cite{savva2019habitat} platform. The underlying architecture of our setup is built on L3MVN~\cite{yu2023l3mvn}. We use GPT-2 as the LLM for inferring spatial semantic information. This study is implemented based on the PyTorch framework, running on the Ubuntu 22.04 operating system and equipped with one NVIDIA A40 graphics card.

\subsection{Metrics}
We evaluate all methods using Success Rate (SR, \%) and Success weighted by Path Length (SPL, \%)~\cite{anderson2018evaluation}. 

\subsection{Baselines}
\begin{itemize}
\item L3MVN~\cite{yu2023l3mvn}: Uses LLMs to infer semantic relevance and selects target-related frontier points as long-term goals.
\item SemUtil~\cite{chen2023not}: Integrates linguistic priors and scene-statistical semantics into geometric frontier exploration for optimal direction selection.
\item VoroNav~\cite{wu2024voronav}: Generates simplified paths via Voronoi diagrams and employs LLMs with hierarchical rewards for medium-term goal selection.
\item ESC~\cite{zhou2023esc}: Combines VLM-based scene detection with LLM commonsense reasoning, incorporating knowledge as soft constraints for navigation.
\item SemExp~\cite{chaplot2020object}: Constructs semantic maps and trains goal-oriented exploration strategies using reinforcement learning for long-term goal selection.
\item PONI~\cite{ramakrishnan2022poni}: Employs potential function networks to predict regional and object potentials for exploration direction, requiring no environment interaction.
\end{itemize}

\subsection{Result and Discussion}
As shown in Table~\ref{tab:results}, HRO achieves a SR of 53.0\% on HM3D and 84.0\% on Gibson, significantly outperforming all baseline methods. Notably, HRO also maintains competitive SPL metrics: 24.5\% on HM3D and 46.5\% on Gibson. This indicates that our method achieves a higher SR without sacrificing path efficiency. The process of finding a chair is illustrated in Fig.~\ref{fig4}.

\begin{table}[t]
\centering
\caption{Performance comparison of different LLM-based ZS-OGN methods on HM3D and Gibson datasets.}
\label{tab:results}
\begin{tabular*}{\linewidth}{@{\extracolsep{\fill}}lccccc}
\toprule
\multirow{2}{*}{Method} & \multirow{2}{*}{Zero-shot} & \multicolumn{2}{c}{HM3D} & \multicolumn{2}{c}{Gibson} \\
\cmidrule(lr){3-4} \cmidrule(lr){5-6}
 & & SR $\uparrow$ & SPL $\uparrow$ & SR $\uparrow$ & SPL $\uparrow$ \\
\midrule
SemExp & - & 37.9 & 18.8 & 65.2 & 33.6 \\
PONI & - & - & - & 73.6 & 41.0 \\
SemUtil & BERT & - & - & 69.3 & 40.5 \\
ESC & GPT-3.5 & 39.2 & 22.3 & - & - \\
VoroNav & GPT-3.5 & 42.0 & \textbf{26.0} & - & - \\
L3MVN & GPT-2 & 50.4 & 23.1 & 76.1 & 37.7 \\
HRO (Ours) & GPT-2 & \textbf{53.0} & 24.5 & \textbf{84.0} & \textbf{46.5} \\
\bottomrule
\end{tabular*}
\end{table}

Compared with SemExp~\cite{chaplot2020object} and PONI~\cite{ramakrishnan2022poni}, SemExp relies on reinforcement learning to train semantic exploration strategies, while PONI predicts regional values through a potential function network. Both methods require specialized training in specific scenarios. In contrast, zero-shot methods have achieved performance close to or even exceeding the evaluation metrics of these two methods, demonstrating that zero-shot methods can maintain better generalization capabilities and possess significant advantages in task versatility. This confirms that the prior knowledge embedded in LLMs can play an effective guiding role in the navigation process.

When comparing with SemUtil~\cite{chen2023not}, VoroNav~\cite{wu2024voronav}, and ESC~\cite{zhou2023esc}—methods that attempt to integrate semantic and geometric information—SemUtil relies on scene-statistical semantics, VoroNav simplifies paths based on Voronoi diagrams, and ESC depends on pre-trained VLMs. Despite SemUtil’s use of BERT embeddings, its performance remains low. On the Gibson dataset, HRO outperforms SemUtil by 14.7\% in SR and 6.0\% in SPL, indicating that our targeted modeling of room-object affinity is more effective than general semantic utility learning. While VoroNav and ESC, which are based on GPT-3.5, achieve decent performance, HRO—built on GPT-2—surpasses them in SR on the HM3D dataset by 11.0\% and 13.8\%, respectively. HRO’s advantages validate that directly leveraging the commonsense reasoning capabilities of LLMs, rather than relying on pre-trained models or scene statistics, enables more accurate capture of ``object to room'' semantic associations in zero-shot scenarios. This also suggests that a stronger language model alone is insufficient; an appropriate reasoning structure is needed to fully unleash the potential of language models.

\begin{table}[t]
\centering
\caption{HRO ablation study results on HM3D dataset.}
\label{tab:ablation}
\begin{tabular*}{\linewidth}{@{\extracolsep{\fill}}cccccc}
\toprule
\multicolumn{4}{c}{HRO ablation} & \multicolumn{2}{c}{HM3D results} \\
\cmidrule(lr){1-4} \cmidrule(lr){5-6}
Object & RT Inf & SA Sco & GT Seg & Success $\uparrow$ & SPL $\uparrow$ \\
\midrule
$\checkmark$ & & & & 50.2 & 23.6 \\
& $\checkmark$ & & & 51.5 & 23.7 \\
& $\checkmark$ & $\checkmark$ & & 53.0 & 24.5 \\
& $\checkmark$ & $\checkmark$ & $\checkmark$ & 64.9 & 35.8 \\
\bottomrule
\end{tabular*}
\end{table}

\begin{figure}[t]
\centering
\includegraphics[width=0.95\linewidth]{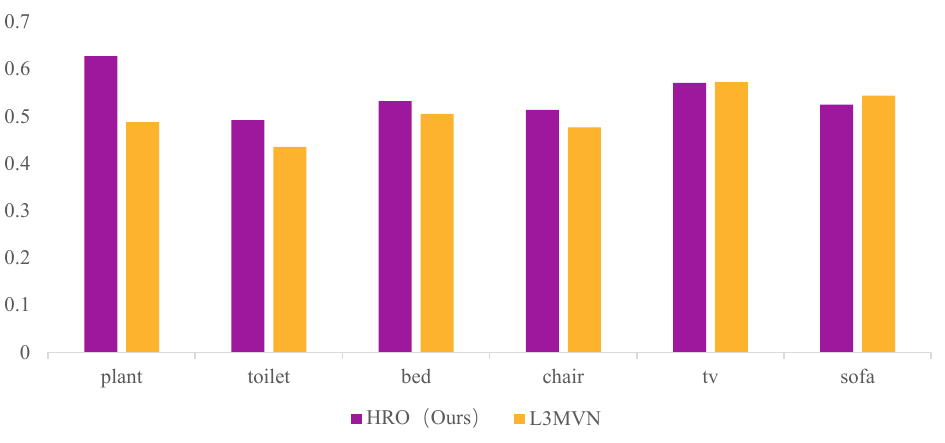}
\caption{Comparison of the success rate of each goal category on HM3D between HRO and L3MVN.
} \label{fig3}
\end{figure}

Fig.~\ref{fig3} illustrates a category-wise comparison between HRO and L3MVN on the HM3D dataset. In comparison with L3MVN~\cite{yu2023l3mvn}, L3MVN only uses LLMs to infer semantic relevance for selecting frontier points, whereas HRO constructs a hierarchical ``room-to-object'' reasoning framework: first performing coarse-grained semantic modeling of room types via LLMs, then refining to region selection for the target object. Both methods use GPT-2 for semantic reasoning, but HRO’s hierarchical room-to-object approach provides more structured semantic reasoning than L3MVN’s direct object-object associations. This layered decision-making enhances the structure of semantic reasoning, leading to significant improvements in both SR and SPL compared to L3MVN.

\subsection{Ablations}
As shown in Table~\ref{tab:ablation}, we quantitatively analyzed the impact of each component on navigation performance by gradually introducing core modules. The object-object association adopts a strategy similar to L3MVN, namely ``direct reasoning based on object semantic relevance''. It selects exploration targets directly through semantic associations between objects, serving as a baseline method without hierarchical semantic reasoning. Room Type Inference (RT Inf) is the high-level module of HRO, which performs semantic reasoning on room types via an LLM. Semantic Affinity Scoring (SA Sco) is the mid-level module of HRO, referring to the target-oriented room affinity scoring. Ground Truth Segmentation (GT Seg) uses the real semantic segmentation results.

The object-object association achieved a SR of 50.2\% and an SPL of 23.6\% (Row 1, Table~\ref{tab:ablation}), validating LLM semantic reasoning as a baseline. When we replaced the direct object association with the core room type inference module proposed in this paper (Row 2, Table~\ref{tab:ablation}), the performance improved significantly, with the SR increasing by 1.3\%. This indicates that upgrading the ``object with object'' flat reasoning to the ``object to room'' hierarchical reasoning can more effectively utilize spatial semantic priors and guide the agent to move toward functional regions with greater potential. On the basis of room type inference, after introducing the semantic affinity scoring module (Row 3, Table~\ref{tab:ablation}), our complete HRO model achieved the best performance, with a SR of 53.0\% and an SPL of 24.5\%. This module successfully realizes the key transformation from ``generalized room semantic understanding'' to ``target-specific navigation decision-making'', closing the loop of the reasoning chain. When ground truth segmentation was integrated into the complete model (Row 4, Table~\ref{tab:ablation}), the performance experienced a substantial leap. This suggests that part of the performance bottleneck of the current method stems from the accuracy of upstream object detection. This is mainly because accurate semantic segmentation significantly improves the quality of semantic map construction, and the inaccuracy of semantic map construction is a key factor leading to task failure.


\section{Conclusion}
This paper proposes HRO, a hierarchical ``room-to-object'' framework for zero-shot object navigation. With room type as the intermediate semantic representation, its coarse-to-fine reasoning mimics human ``scene-first, object-second'' exploration and improves navigation efficiency. Future work will focus on deeper language-environment interaction mechanisms to further unleash LLMs’ potential in navigation.

\addtolength{\textheight}{-12cm}   









\bibliographystyle{IEEEtran}   
\bibliography{mybibliography}

\end{document}